\relax
\documentclass[11pt]{article}
\usepackage[textwidth=400pt,centering]{geometry}
\usepackage{fullpage}
\usepackage{microtype}
\usepackage{graphicx}
\usepackage{subfigure}
\usepackage{booktabs} 
\usepackage{xcolor}
\usepackage[authoryear]{natbib}
\usepackage[nottoc,notlot,notlof]{tocbibind}
\usepackage[colorlinks=true,linkcolor=blue, citecolor=blue]{hyperref}
\usepackage{url}            
\usepackage{amsmath}
\usepackage{times}
\usepackage{graphicx}
\usepackage{color}
\usepackage{multirow}
\usepackage{rotating}
\usepackage{bbm}
\usepackage{latexsym}
\usepackage{amssymb}
\usepackage{booktabs}
\usepackage{xcolor}
\usepackage{bm}
\usepackage{amsthm}

\newtheorem{lemm}{Lemma}
\newtheorem{theo}{Theorem}

\newcommand{\xs}{x_a, x_b, x_c}
\newcommand{\ys}{y_a, y_b, y_c}
\newcommand{\pip}{\pi_+}
\newcommand{\pin}{\pi_-}
\newcommand{\pit}{\pi_{\mathrm{T}}}
\newcommand{\pite}{\pi_{\mathrm{test}}}
\newcommand{\pp}{p_+}
\newcommand{\pn}{p_-}
\newcommand{\Ep}{\mathop{\mathbb{E}}}

\newcommand{\lfp}{\ell(f(x), +1)}
\newcommand{\lfn}{\ell(f(x), -1)}

\newcommand{\argmin}{\mathop{\rm arg~min}\limits}

\begin{document}
\title{Classification from Triplet Comparison Data}
\author{Zhenghang Cui$^{1,2}$ \and Nontawat Charoenphakdee$^{1,2}$  \and Issei Sato$^{1,2}$ \and Masashi Sugiyama$^{2,1}$}
\date{
$^1$ The University of Tokyo 
$^2$ RIKEN AIP
}
\maketitle

\begin{abstract}
Learning from triplet comparison data has been extensively studied in the context of metric learning, where we want to learn a distance metric between two instances, and ordinal embedding, where we want to learn an embedding in an Euclidean space of the given instances that preserves the comparison order as well as possible.
Unlike fully-labeled data, triplet comparison data can be collected in a more accurate and human-friendly way.
Although learning from triplet comparison data has been considered in many applications, an important fundamental question of whether we can learn a classifier \textit{only} from triplet comparison data has remained unanswered. 
In this paper, we give a \textit{positive} answer to this important question by proposing an \textit{unbiased estimator} for the \textit{classification risk} under the empirical risk minimization framework.
Since the proposed method is based on the empirical risk minimization framework, it inherently has the advantage that any surrogate loss function and any model, including \textit{neural networks}, can be easily applied.
Furthermore, we theoretically establish an estimation error bound for the proposed empirical risk minimizer.
Finally, we provide experimental results to show that our method empirically works well and outperforms various baseline methods.
\end{abstract}
\section{Introduction}
Recently, learning from comparison-feedback data has received increasing attention~\citep{Heim16, Kleindessner17}.
It is usually argued that humans perform better in the task of evaluating which instances are similar, rather than identifying each individual instance~\citep{Stewart05}.
It is also argued that humans can achieve much better and more reliable performance on assessing the similarity on a relative scale (``Instance A is more similar to instance B than to instance C") rather than on an absolute scale (``The similarity score between A and B is $0.9$ while the one between A and C is $0.4$")~\citep{Kleindessner17}.
Collecting data in this manner has the advantage of avoiding the problem caused by individuals' different assessment scales.
On the other hand, the collected absolute similarity scores may only provide information on a comparison level in some applications, e.g., sensor localization~\citep{Liu2004}.
It was shown that keeping only the relative comparison information can help an algorithm be resilient against measurement errors and achieve high accuracy~\citep{Xiao06}.

In this paper, we focus on the problem of learning from triplet comparison data, which is a common form of comparison-feedback data.
A triplet comparison $(x_a, x_b, x_c)$ contains the information that instance $x_a$ is more similar to $x_b$ than to $x_c$.
As one example, search-engine query logs can readily provide feedback in the form of triplet comparisons~\citep{Schultz04}.
Given a list of website links $\{A, B, C\}$ for a query, if links $A$ and $B$ are clicked and the link $C$ is not clicked, we can formulate a triplet comparison as $(A, B, C)$.

Learning from triplet comparison data was initially studied in the context of metric learning~\citep{Schultz04}, in which a consistent distance metric between two instances is assumed to be learned from data.
The well-known triplet loss for face recognition was proposed in this line of research~\citep{Schroff15, Yu18}.
Using this loss function, an inductive mapping function can be efficiently learned from triplet comparison image data.
At the same time, the problem of ordinal embedding has also been extensively studied~\citep{Agarwal07, Van12}. It aims to learn an embedding of the given instances to the Euclidean space that preserves the order given by the data.
Algorithms for large scale ordinal embedding have been developed~\citep{Anderton19}.
In addition, many other problem settings have been considered for the situation of using only triplet comparison data,
such as nearest neighbor search~\citep{Haghiri17}, kernel function construction~\citep{KleindessnerKernel} and outlier identification ~\citep{KleindessnerLens}.

However, learning a binary classifier from triplet comparison data remained untouched until recently.
A random forest construction algorithm~\citep{Haghiri18} was proposed for both classification and regression.
However, it first requires a labeled dataset and needs to \textit{actively} access a triplet comparison oracle many times.
For \textit{passively} collected triplet comparison data, a boosting based algorithm~\citep{Perrot18} was recently proposed without accessing a triplet comparison oracle. However, a set of labeled data is still indispensable to initiating the training process.
To the best of our knowledge,
this paper is the first to tackle the problem of learning a classifier \textit{only} from \textit{passively} obtained triplet comparison data, without accessing either a labeled dataset or an oracle.

\paragraph{Contributions:}
We show that we can learn a binary classifier from only passively obtained triplet comparison data.
We achieve this goal by developping a novel method for learning a binary classifier in this setting with theoretical justification.
We use the direct risk minimization framework given for the classification problem.
We then show that the classification risk can be empirically estimated in an unbiased way given \textit{only} triplet comparison data.
Theoretically, we establish an estimation error bound for the proposed empirical risk minimizer, showing that learning from triplet comparison data is consistent.
Our method also returns an \textit{inductive} model, which is different from clustering and ordinal embedding, and can be applied to unseen test data points.
The test data would consist of single instances instead of triplet comparisons since our primitive goal is to perform a binary classification task on unseen data points.

In summary, for the problem of classification using \textit{only} triplet comparison data, our contributions in this paper are three-fold:
\begin{itemize}
    \item We propose an empirical risk minimization method for binary classification using \textit{only} passively obtained triplet comparison data, which gives us an inductive classifier.
    \item We theoretically establish an estimation error bound for our method, showing that the learning is consistent.
    \item We experimentally demonstrate the practical usefulness of our method.
\end{itemize}

\section{Related Work}
Our problem setting of learning a binary classifier from \textit{passively} obtained triplet comparison data can be considered as a type of a weakly-supervised classification problem, where we do not have access to ground-truth labels~\citep{Zhou17}.

An approach based on constructing an unbiased risk estimator of the true classification risk from weakly-supervised data has been explored in many problem settings; for example, positive-unlabeled classification~\citep{du14,Niu16} and similarity-unlabeled classification~\citep{Bao18} can be handled by the framework of learning from two sets of unlabeled data~\citep{Lu18}.
Nevertheless, our problem setting is not a special case addressed by~\citet{Lu18} since we have only one set of triplet comparison data.
We later show that we can formulate three different distributions, which is fundamentally different from the framework used by ~\cite{Lu18}.

Moreover, our problem setting is also different from similarity-dissimilarity-unlabeled classification~\citep{Shimada19} in the sense that we have no access to unlabeled data and similarity and dissimilarity pairs, but \emph{only triplet comparison information}.
Furthermore, it is important to note that our problem setting is also different from preference learning~\citep{Furnkranz10}, since we do not want to learn a ranking function but construct a binary classifier.
Although we can first learn a ranking function and then decide a proper threshold to construct a binary classifier~\citep{Narasimhan13},
it is not straightforward to choose a proper threshold.
Therefore, instead of this two-stage method, we focus on a method that can directly learn a binary classifier from triplet comparison data.

\section{Learning A Classifier from Triplet Comparison Data}
In this section,
we first review the ordinary fully supervised classification setting.
Then we introduce the problem setting and assumption for the data generation process of triplet comparison data.
Finally, we describe the proposed method for training a binary classifier from only passively obtained triplet comparison data.

\subsection{Preliminary}
We first briefly introduce the traditional binary classification problem.
We denote $\mathcal{X}\subset\mathbb{R}^d$ as a $d$-dimensional sample space and $\mathcal{Y} = \{+1, -1\}$ as a binary label space.
In the fully supervised setting,
we usually assume the labeled data $(x,y)\in\mathcal{X}\times\mathcal{Y}$ are drawn from the joint probability distribution with density $p(x, y)$~\citep{Vapnik95}.
The goal is to obtain a classifier $f:\mathcal{X}\rightarrow\mathbb{R}$ that minimizes the classification risk
\begin{equation}
  R(f) = \Ep_{(x,y)\sim p(x,y)}[\ell(f(x), y)],
\end{equation}
where the expectation is over the joint density $p(x,y)$ and $\ell:\mathbb{R}\times\mathcal{Y}\rightarrow\mathbb{R}_+$ is a loss function that measures how well the classifier estimates the true class label.

In the traditional fully supervised classification setting,
we are given both positive and negative training data collectively drawn from the joint density $p(x, y)$.
However, in our case,
we still want to train a binary classifier that minimizes the classification risk, although we do not have fully labeled data.

\subsection{Generation Process of Triplet Comparison Data}
\label{section:generation}
We formulate the underlying generation process of triplet comparison data in order to perform empirical risk minimization.
Three samples in a triplet are first generated independently, then shown to a user.
The user can mark the triplet to be proper or not.
A proper triplet means that the similarity between the first and second samples is stronger or the same as the similarity between the first and third samples.
Specifically, it means that three labels $(\ys)$ in a triplet appear to be one of the following cases:
\begin{equation*}
\begin{split}
  \mathcal{Y}_1\triangleq\{& (+1, +1, -1), (-1, -1, +1), (+1, +1, +1),\\
  & (-1, -1, -1), (+1, -1, -1), (-1, +1, +1)\}.
\end{split}
\end{equation*}
Otherwise, it means the first sample is more similar to the third sample than to the second sample; thus, the user chooses to mark the triplet as not proper.
Similarly, it means $(\ys)$ appears to be one of the following cases $$\mathcal{Y}_2\triangleq\{(+1, -1, +1), (-1, +1, -1)\}.$$

First, three data samples are generated independently from the underlying joint density $p(x, y)$, then $\mathcal{D} = \{(\xs)\}$ are collected without knowing the underlying true labels $(\ys)$.
However, we can collect information about which case a triplet belongs to from user feedback.
After receiving feedback from users,
we can actually obtain two distinct datasets.
The data the user chooses to keep the order is denoted as $$\mathcal{D}_1\triangleq\{(\xs)|(\ys)\in\mathcal{Y}_1\}.$$
Similarly, the data the user chooses to flip the order is denoted as $$\mathcal{D}_2\triangleq\{(\xs)|(\ys)\in\mathcal{Y}_2\}.$$

Note that the ratio of $n_1\triangleq|\mathcal{D}_1|$ to $n_2\triangleq|\mathcal{D}_2|$ is fixed because we assume the three samples in a triplet are generated independently from $p(x,y)$; thus, the ratio $\frac{n_1}{n_2}$ is only dependent on the underlying class prior probabilities, which are fixed unknown values.

The two datasets can be considered to be generated from two underlying distributions as indicated by the following lemma.

\begin{lemm}
Corresponding to the data generation process described above, let
\begin{equation}
\begin{split}
  p_1(\xs) & = \frac{p(\xs,(\ys)\in\mathcal{Y}_1)}{\pit}, \\
  p_2(\xs) & = \pip\pp(x_a)\pn(x_b)\pp(x_c) + \pin\pn(x_a)\pp(x_b)\pn(x_c), \\
\end{split}
\end{equation}
where $\pit=1-\pip\pin$, $\pip\triangleq p(y=+1)$ and $\pin\triangleq p(y=-1)$ are the class prior probabilities that satisfy $\pip + \pin = 1$ and
$\pp(x)\triangleq p(x|y=+1)$ and $\pn(x)\triangleq p(x|y=-1)$ are class conditional probabilities.
Then it follows
\begin{equation*}
\begin{split}
  \mathcal{D}_1 = \{(x_{1,a}, x_{1,b}, x_{1,c})\}_{i=1}^{n_1} \mathop{\sim}^{\mathrm{i.i.d.}} p_1(\xs), \\
  \mathcal{D}_2 = \{(x_{2,a}, x_{2, b}, x_{2, c})\}_{i=1}^{n_2} \mathop{\sim}^{\mathrm{i.i.d.}} p_2(\xs). \\
\end{split}
\end{equation*}
\end{lemm}

Detailed derivation is given in Appendix~\ref{proof:lemma1}.

We denote the pointwise data collected from $\mathcal{D}_1$ and $\mathcal{D}_2$ by ignoring the triplet comparison relation as $\mathcal{D}_{1,a}\triangleq\{x_{1,a}\}_{i=1}^{n_1}$, $\mathcal{D}_{1,b}\triangleq\{x_{1,b}\}_{i=1}^{n_1}$, $\mathcal{D}_{1,c}\triangleq\{x_{1,c}\}_{i=1}^{n_1}$, $\mathcal{D}_{2,a}\triangleq\{x_{2,a}\}_{i=1}^{n_2}$, $\mathcal{D}_{2,b}\triangleq\{x_{2,b}\}_{i=1}^{n_2}$ and $\mathcal{D}_{2,c}\triangleq\{x_{2,c}\}_{i=1}^{n_2}$,
the marginal densities of which can be expressed by the following theorem.

\begin{theo}
  \label{theo:pointwise}
  Samples in $\mathcal{D}_{1,a}$, $\mathcal{D}_{1,c}$, $\mathcal{D}_{2,a}$ and $\mathcal{D}_{2,c}$ are independently drawn from
  \begin{equation}
    \tilde p_1(x) = \pip\pp(x) + \pin\pn(x),
  \end{equation}
  samples in $\mathcal{D}_{1,b}$ are independently drawn from
  \begin{equation}
    \tilde p_2(x) = \frac{(\pip^3+2\pip^2\pin)\pp(x) + (2\pip\pin^2+\pin^3)\pn(x)}{\pit},
  \end{equation}
  and samples in $\mathcal{D}_{2,b}$ are independently drawn from
  \begin{equation}
    \tilde p_3(x) = \pin\pp(x) + \pip\pn(x).
  \end{equation}
\end{theo}

A proof is given in Appendix~\ref{proof:1}.

Theorem \ref{theo:pointwise} indicates that from triplet comparison data, we can essentially obtain samples that can be drawn independently from three different distributions.
We denote the three aggregated datasets as
\begin{equation*}
  \begin{split}
    \tilde{\mathcal{D}}_1 & = \mathcal{D}_{1,a} \cup \mathcal{D}_{1,c} \cup \mathcal{D}_{2,a} \cup \mathcal{D}_{2,c}, \\
    \tilde{\mathcal{D}}_2 & = \mathcal{D}_{1,b}, \quad
    \tilde{\mathcal{D}}_3 = \mathcal{D}_{2,b}.
  \end{split}
\end{equation*}

\subsection{Unbiased Risk Estimator for Triplet Comparison Data}
We now attempt to express the classification risk,
\begin{equation}
 R(f) \triangleq \Ep_{(x,y)\sim p(x,y)}[\ell(f(x), y)],
\end{equation}
on the basis of the three pointwise densities presented in Section~\ref{section:generation}.

The classification risk can be separately expressed as the expectations over $\pp(x)$ and $\pn(x)$.
Although we do not have access to data drawn from these two distributions,
we can obtain data from three related densities $\tilde p_1(x)$, $\tilde p_2(x)$, and $\tilde p_3(x)$ as indicated in Theorem \ref{theo:pointwise}.
Letting
  \begin{equation}
    A \triangleq \frac{\pip^3+2\pip^2\pin}{\pit},\quad
    B \triangleq \frac{2\pip\pin^2+\pin^3}{\pit},
  \end{equation}
we can express the relationship between these densities as
\begin{equation}
  \label{eq:before-inverse}
  \begin{bmatrix} \tilde p_1(x) \\ \tilde p_2(x) \\ \tilde p_3(x)
  \end{bmatrix}
  = 
  \begin{bmatrix}
  \pip & \pin \\ A & B \\ \pin & \pip
  \end{bmatrix}
  \begin{bmatrix}
  \pp(x) \\ \pn(x)
  \end{bmatrix}.
\end{equation}

Our goal is to solve the above equation so that we can express $\pp(x)$ and $\pn(x)$ in terms of the three densities from which we have i.i.d. data samples.
To this end, we can rewrite the classification risk, which we want to minimize, in terms of $\tilde p_1(x)$, $\tilde p_2(x)$ and $\tilde p_3(x)$.
An answer to Eq.\,\eqref{eq:before-inverse} is given by the following lemma.

\begin{lemm}
We can express $\pp(x)$ and $\pn(x)$ in terms of $\tilde p_1(x)$, $\tilde p_2(x)$ and $\tilde p_3(x)$ as
\begin{equation}
\label{eq:rewrite}
\begin{split}
  \pp(x) & = \frac1{(ac-b^2)}\left((c\pip-b\pin)\tilde p_1(x)+(cA-bB)\tilde p_2(x)+(c\pin-b\pip)\tilde p_3(x)\right), \\
  \pn(x) & = \frac1{(ac-b^2)}\left((a\pin-b\pip)\tilde p_1(x)+(aB-bA)\tilde p_2(x)+(a\pip-b\pin)\tilde p_3(x)\right), \\
\end{split}
\end{equation}
provided $ac-b^2\neq0$ where
\begin{equation*}
  a \triangleq \pip^2+A^2+\pin^2,\quad
  b \triangleq 2\pip\pin+AB,\quad
  c \triangleq \pin^2+B^2+\pip^2.
\end{equation*}
\end{lemm}

Detailed derivation is given in Appendix~\ref{proof:lemma2}.

As a result of the above lemma,
we can express the classification risk using only triplet comparison data.
Letting $\ell_+(x)\triangleq\ell(f(x),+1)$ and $\ell_-(x)\triangleq\ell(f(x),-1)$, we have the following theorem.

\begin{theo}
  The classification risk can be equivalently expressed as
  \begin{equation}
    \label{eq:theorem2}
    \begin{split}
      R(f) = \frac1{(ac-b^2)}\{
      & \Ep_{x\sim \tilde p_1(x)}\left[\pite(c\pip-b\pin)\,\ell_+(x)+(1-\pite)(a\pin-b\pip)\,\ell_-(x)\right] + \\
      & \Ep_{x\sim \tilde p_2(x)}\left[\pite(cA-bB)\,\ell_+(x)+(1-\pite)(aB-bA)\,\ell_-(x)\right] + \\
      & \Ep_{x\sim \tilde p_3(x)}\left[\pite(c\pin-b\pip)\,\ell_+(x)+(1-\pite)(a\pip-b\pin)\,\ell_-(x)\right] \},
    \end{split}
  \end{equation}
  where $\pite\triangleq p_\mathrm{test}(y=+1)$ denotes the class prior of the test dataset.
\end{theo}

A proof is given in Appendix~\ref{proof:2}.

In this paper, we consider the common case in which $\pite=\pip$, which means the test dataset shares the same class prior as the training dataset.
However, even when $\pite\neq\pip$, which means the class prior shift~\citep{Sugiyama12} occurs, our method can still be used when $\pite$ is known.

\section{Estimation Error Bound}
In this section, we establish an estimation error bound for the proposed unbiased risk estimator.
Let $\mathcal{F}\subset\mathbb{R}^{\mathcal{X}}$ represent a function class specified by a model.
First, let $\mathfrak{R}(\mathcal{F})$ be the (expected) Rademacher complexity of $\mathcal{F}$ which is defined as
\begin{equation}
  \mathfrak{R}(\mathcal{F})\triangleq\Ep_{Z_1,\cdots,Z_n\sim\mu}\Ep_{\bm{\sigma}}\left[\sup_{f\in\mathcal{F}}\frac1n\sum_{i=1}^n\sigma_if(Z_i)\right]
\end{equation}
where $n$ is a positive integer,
$Z_1,\cdots,Z_n$ are i.i.d. random variables drawn from a probability distribution with density $\mu$, and $\bm{\sigma} = (\sigma_1,\cdots,\sigma_n)$ are Rademacher variables, which are random variables that take the value of $+1$ or $-1$ with even probabilities.

We assume for any probability density $\mu$,
the specified model $\mathcal{F}$ satisfies
$\mathfrak{R}(\mathcal{F})\leq\frac{C_\mathcal{F}}{\sqrt{n}}$
for some constant $C_\mathcal{F}>0$.
Also let $f^*\triangleq\argmin_{f\in\mathcal{F}} R(f)$ be the true risk minimizer and $\hat f\triangleq\argmin_{f\in\mathcal{F}} \hat R_{T,\ell}(f)$ be the empirical risk minimizer.

\begin{theo}
Assume the loss function $\ell$ is $\rho$-Lipschitz with respect to the first argument ($0<\rho<\infty$), and all functions in the model class $\mathcal{F}$ are bounded, i.e., there exists a constant $C_b$ such that $||f||_\infty\leq C_b$ for any $f\in\mathcal{F}$. Let $C_\ell\triangleq\sup_{t\in\{\pm1\}}\ell(C_b,t)$. Then for any $\delta>0$, with probability at least $1-\delta$:
\begin{equation}
  R(\hat{f}) - R(f^*) \leq \left( \frac{2\rho C_\mathcal{F}}{\sqrt{n}} + \sqrt{\frac{C_\ell^2\log\frac2\delta}{2n}}\right) \cdot
  \frac{C_R}{|ac-b^2|},
\end{equation}
where
\begin{equation}
  \begin{split}
    C_R = & |\pite(c\pip-b\pin)| + |(1-\pite)(a\pin-b\pip)| + |\pite(cA-bB)| + \\ & |(1-\pite)(aB-bA)| + |\pite(c\pin-b\pip)| + |(1-\pite)(a\pip-b\pin)|.
  \end{split}
\end{equation}
\end{theo}

A proof is given in Appendix~\ref{proof:3}.

\begin{figure}
\begin{center}
\includegraphics[width=0.7\textwidth]{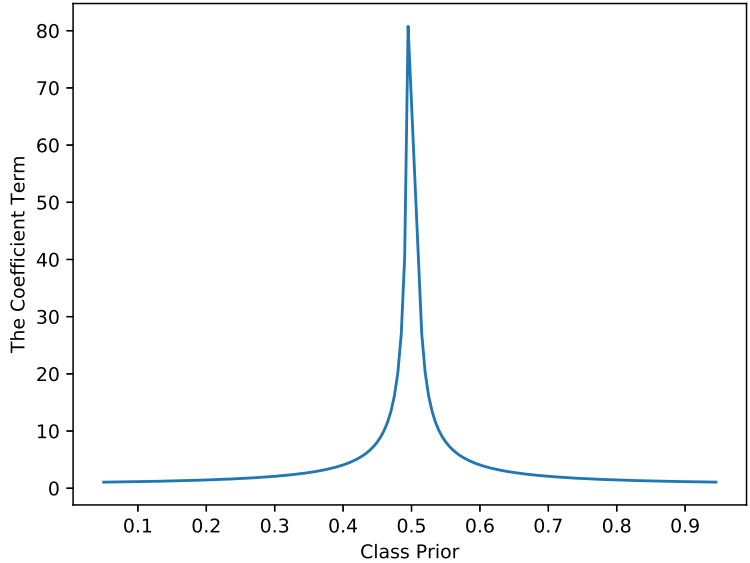}
\caption{Behaviour of the coefficient term.}
\label{fig:bound}
\end{center}
\end{figure}

Since $n$ appears in the denominator, it is obvious that when the class prior is fixed,
the bound will get tighter as the amount of triplet comparison data increases.
However, it is not clear how the bound will behave when we fix the amount of triplet comparison data and change the class prior.
Thus in Figure~\ref{fig:bound},
we show the behavior of the coefficient term $\frac{C_R}{|ac-b^2|}$ with respect to the same class prior of both training and test datasets.
From the illustration,
we can capture the rough trend that the bound gets tighter when the class prior becomes further from $0.5$.
We will further investigate this behavior in experiments.

\section{On Class Prior}
In the previous sections, the class prior $\pi_+$ is assumed known.
For this simple case,
we can directly use the proposed algorithm to
separate test data as well as identify correct classes.
However, it may not be true for many real-world applications.
There are two situations that can be considered.
For the worst case,
no information about the class prior is given.
Although we still can estimate a result for the class prior from data and obtain a classifier that is able to separate data for different classes,
we cannot identify the correct class without the information of which class has a higher class prior.
A better situation is that we have the information of which class has a higher class prior.
By setting this class as the positive class,
we can successfully train a classifier to identify the correct class.
Thus, we assume that the positive class has a higher class prior, which means $\pip>\frac{1}{2}$.

\subsection{Class Prior Estimation from Triplet Comparison Data}
Noticing $\pit=1-\pip+\pip^2$,
we can obtain $\pip^2 - \pip + (1-\pit)=0$.
By assuming $\pip>\pin$,
we have
\begin{equation}
  \pip=\frac{1+\sqrt{1-4(1-\pit)}}{2}.
\end{equation}
Since we can unbiasedly estimate $\pit$ by $\frac{n_1}{n_1+n_2}$,
the class prior $\pip$ can thus be estimated once the triplet comparison dataset is given.

\section{Experiments}
In this section, we conducted experiments using real world datasets to evaluate and investigate the performance of the proposed method for triplet classification.

\subsection{Baseline methods}
\paragraph{KMEANS:}
As a simple baseline, we used $k$-means clustering~\citep{Macqueen67} with $k=2$ on all the data instances of triplets while ignoring all the relation information.
\paragraph{ITML:}
Information-theoretic metric learning~\citep{Davis07} is a metric learning method that requires pairwise the relationship between data instances.
From a triplet $(x_a,x_b,x_c)$,
we constructed pairwise constraints as $(x_a,x_b)$ being similar and $(x_a,x_c)$ being dissimilar.
Using the metric returned by the algorithm,
we conducted $k$-means clustering on test data.
We used the identity matrix for prior knowledge and fix the slack variable as $\gamma=1$.
\paragraph{TL:}
Triplet loss~\citep{Schroff15} is a loss function proposed in the context of deep metric learning which can learn a metric directly from triplet comparison data.
Using the metric returned by the algorithm,
we conducted $k$-means clustering on test data.
\paragraph{SERAPH:}
Semi-supervised metric learning paradigm with hyper sparsity~\citep{Niu14} is a metric learning method based on entropy regularization.
We formulated a pairwise relationship in the same manner as with ITML.
Using the metric returned by ITML,
we conducted $k$-means clustering on test data.
\paragraph{SU:} SU learning~\citep{Bao18} is a method for learning a binary classifier from similarity and unlabeled data. We used the same method for estimating the class prior, and considered the less similar sample in a triplet as unlabeled data.

\subsection{Datasets}
\paragraph{UCI datasets:}
We used six datasets from the \textit{UCI Machine Learning Repository}~\citep{Asuncion07}.
They are binary classification datasets and we use the given labels for further triplet comparison data generation.

\paragraph{Image datasets:}
We used the following three image datasets.

The MNIST~\citep{mnist} dataset consists of $70,000$ examples associated with a label from ten digits.
Each data instance is a $28\times28$ gray-scale image; thus, the input dimension is $784$.
To form a binary classification problem,
we treat even numbers as the positive class and odd numbers as the negative class.
The data were standardized to have zero mean and unit variance.

The Fashion MNIST~\citep{fashionmnist} dataset consists of $70,000$ examples associated with a label from ten fashion item classes.
Each data instance is a $28\times28$ gray-scale image thus the input dimension is $784$.
To form a binary classification problem,
we treat five classes, i.e., T-shirt/top, Pullover, Dress, Coat, and Shirt, as positive class since they all represent upper body clothing.
The data were standardized to have zero mean and unit variance.

The CIFAR-10~\citep{cifar10} dataset consists of $60,000$ examples associated with a label from ten classes.
Each image is given in a $32\times32\times3$ format thus the input dimension is $3,072$.
To form a binary classification problem,
we treated four classes, i.e., airplane, automobile, ship, and truck, as positive classe since they all represent artificial objects.

\subsection{Proposed method}
For the proposed method, we used a fully-connected neural network with only $1$ hidden layer of width $100$ and rectified linear units (ReLUs)~\citep{relu} for all the datasets except for CIFAR-10.
The width of the hidden layer was set to be $100$ through out all experiments.
Adam~\citep{Kingma14} was used for optimization.
The neural network architecture used for CIFAR-10 is specified in Appendix.

\subsection{Results}
\begin{figure}
\centering
\begin{minipage}{0.48\hsize}
\includegraphics[width=\columnwidth]{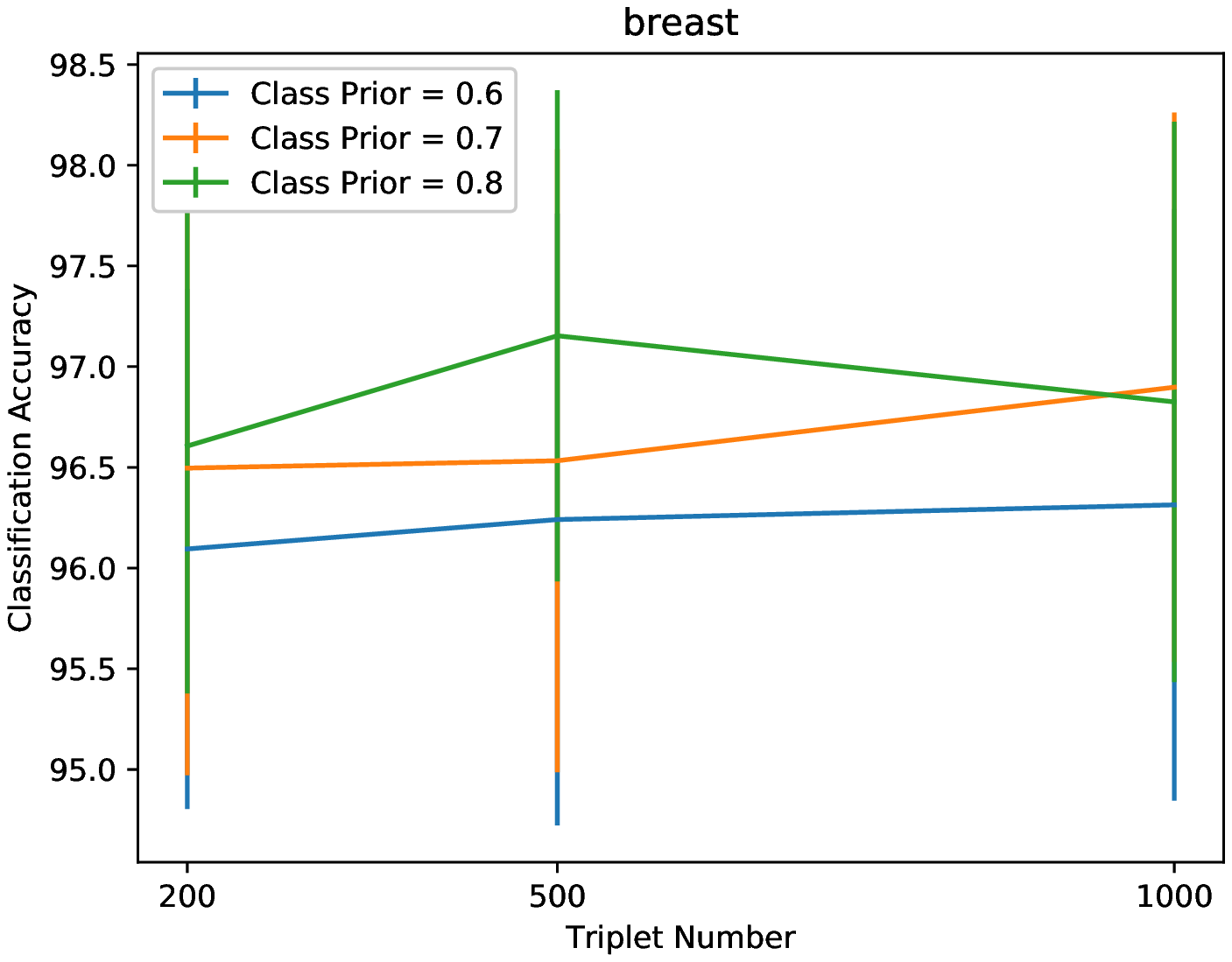}
\end{minipage}
\begin{minipage}{0.48\hsize}
\includegraphics[width=\columnwidth]{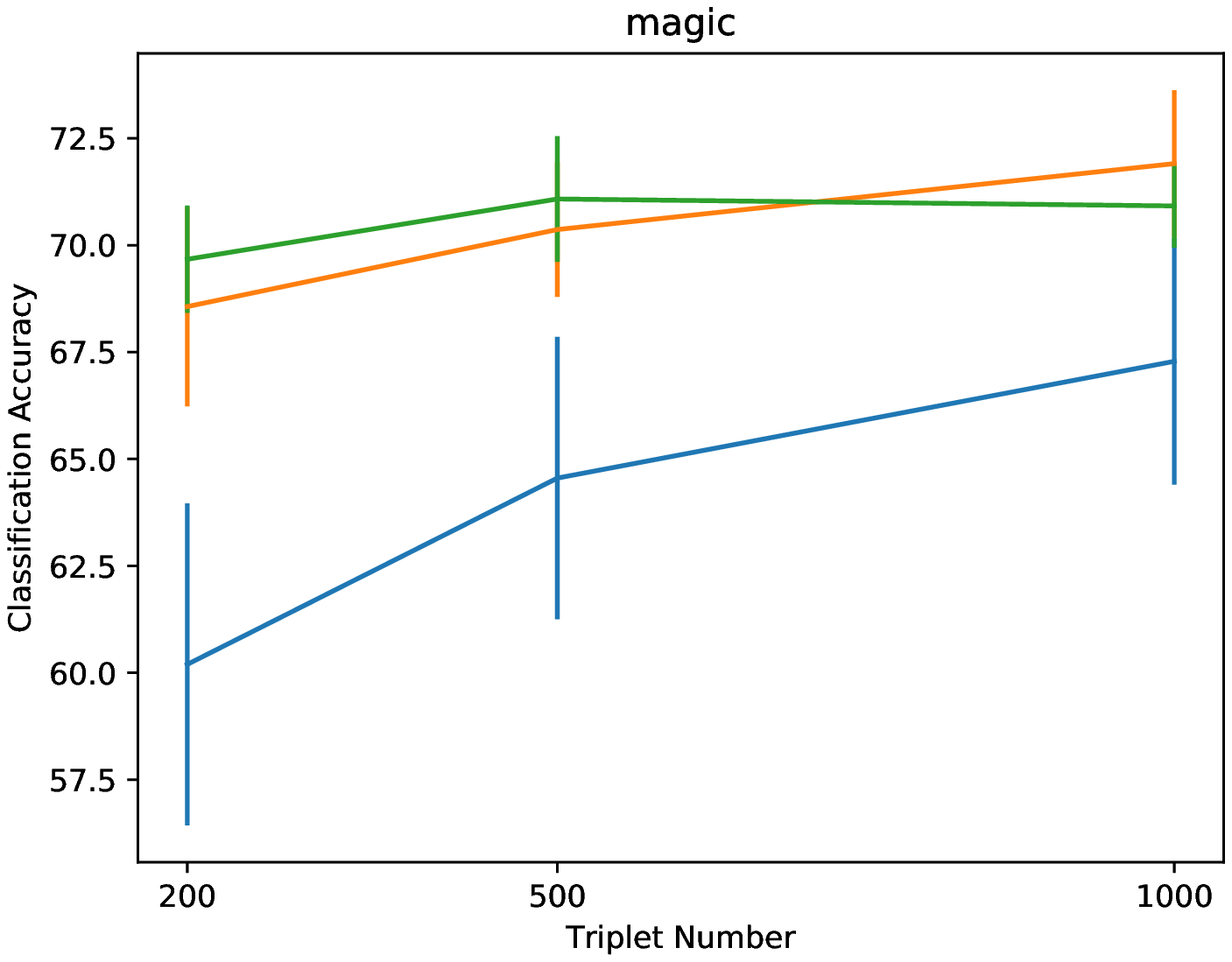}
\end{minipage}
\begin{minipage}{0.48\hsize}
\includegraphics[width=\columnwidth]{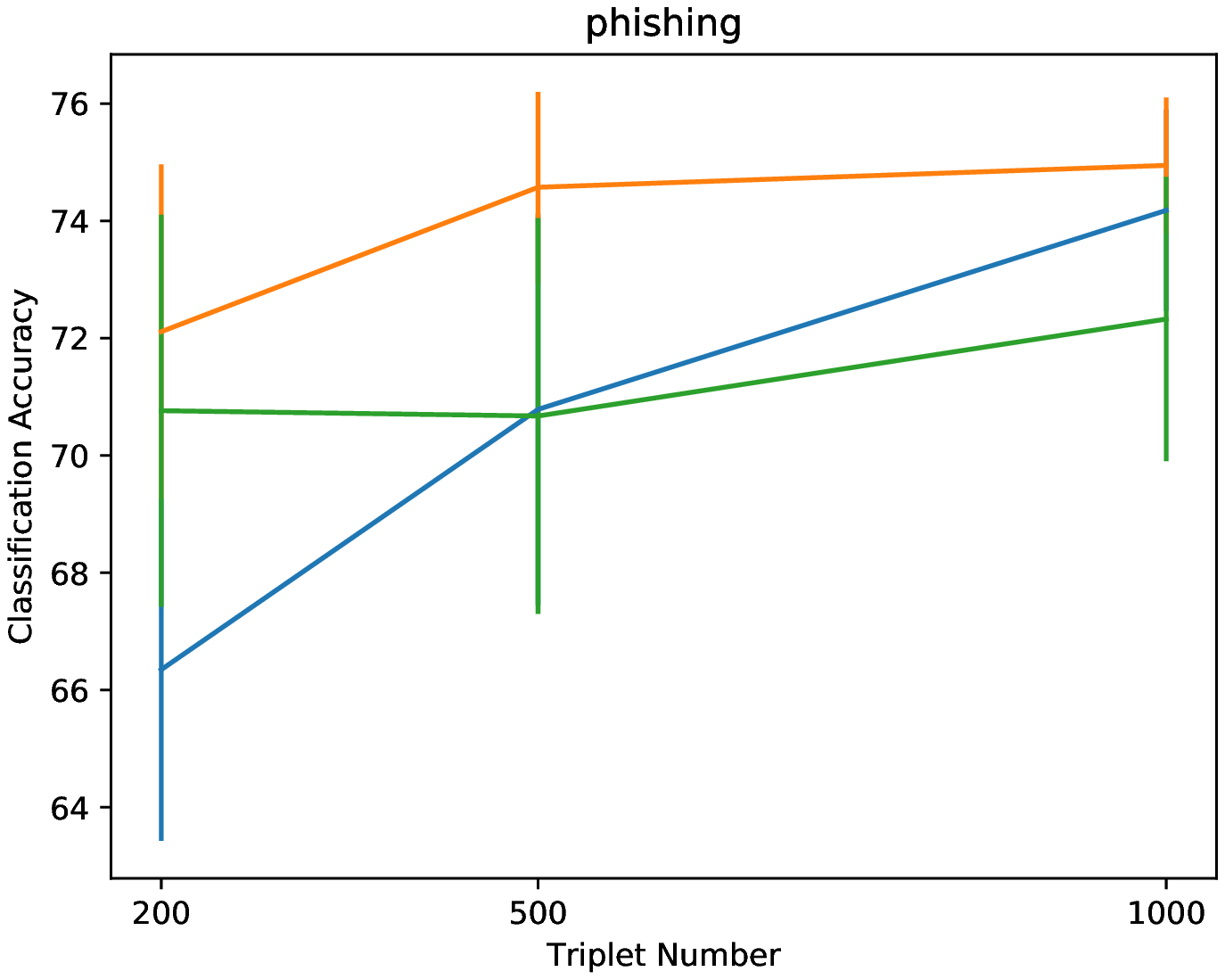}
\end{minipage}
\begin{minipage}{0.48\hsize}
\includegraphics[width=\columnwidth]{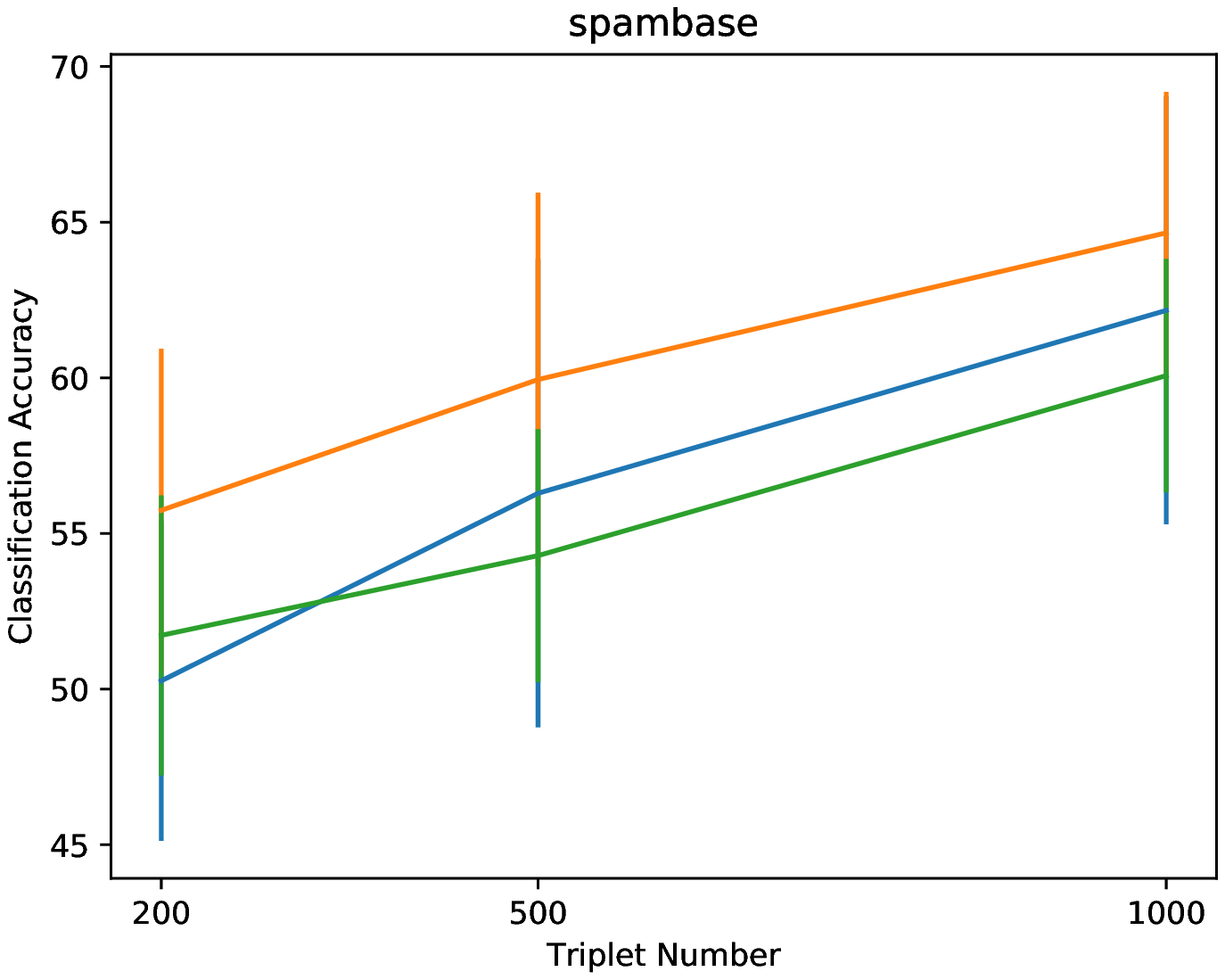}
\end{minipage}
\caption{Average classification error and standard error over $20$ trials.}
\label{fig:exp}
\end{figure}

The proposed method estimates the unknown class prior first.
For baseline methods,
performances are measured by the clustering accuracy $1-\min(r,1-r)$ where $r$ is the error rate.
The results of different triplet numbers are listed in Tables~\ref{table:1}, \ref{table:2}, and \ref{table:3}.
The best and equivalent methods are shown in bold face on the one-sided t-test with a significance level of $5\%$.
Also as shown in Figure~\ref{fig:exp},
the performance of the proposed method with respect to the class prior and the size of training dataset followed the prediction by the theory in most of the cases.

\begin{table}
\centering
\caption{Experimental results with class prior as $0.7$ and $1000$ training triplets.}
\label{table:1}
\scalebox{0.73}{
\begin{tabular} {cccccccc} \hline
& \multicolumn{2}{c}{Proposed Methods} & \multicolumn{5}{c}{Baselines} \\ \cmidrule(lr){2-3} \cmidrule(lr){4-8}
\text{Dataset} & Squared & Double Hinge & KMEANS & ITML & TL & SERAPH & SU\\ \hline
\text{adult} &65.54 (0.41) &64.19 (0.61) &71.94 (0.10) &71.04 (1.00) &61.48 (1.36) &71.04 (1.00) & \textbf{75.88 (0.50)}\\
\text{breast} & \textbf{97.41 (0.28)} & \textbf{96.90 (0.31)} &96.20 (0.34) &95.84 (0.29) &93.87 (0.78) &96.72 (0.23) &65.26 (0.76)\\
\text{diabetes} & \textbf{70.71 (0.84)} &64.87 (0.74) &66.69 (0.70) &65.91 (0.69) &64.38 (1.60) &67.44 (0.78) &34.42 (0.73)\\
\text{magic} &61.75 (1.00) & \textbf{71.91 (0.39)} &65.08 (0.17) &64.79 (0.17) &65.42 (0.22) &64.96 (0.19) &34.77 (0.19)\\
\text{phishing} & \textbf{76.58 (0.30)} &74.95 (0.27) &63.43 (0.50) &63.75 (0.23) &57.85 (0.92) &63.42 (0.53) &34.17 (0.22)\\
\text{spambase} & \textbf{62.08 (1.87)} & \textbf{64.66 (1.04)} & \textbf{63.59 (0.24)} & \textbf{63.24 (0.31)} &59.59 (1.57) & \textbf{63.28 (0.34)} &60.27 (0.30)\\
\text{mnist} &79.86 (0.35) & \textbf{80.78 (0.34)} &65.24 (0.25) &0.00 (0.00) &58.26 (1.24) &0.00 (0.00) &50.80 (0.03)\\
\text{fashion} &89.73 (0.33) & \textbf{91.62 (0.33)} &74.90 (1.00) &0.00 (0.00) &76.83 (1.31) &0.00 (0.00) &49.85 (0.08)\\
\text{cifar10} & \textbf{76.39 (1.57)} &66.28 (2.51) &64.17 (0.01) &0.00 (0.00) &60.17 (1.26) &0.00 (0.00) &59.50 (0.50)\\
\hline
\end{tabular}}
\end{table}

\begin{table}
\centering
\caption{Experimental results with class prior as $0.7$ and $500$ training triplets.}
\label{table:2}
\scalebox{0.73}{
\begin{tabular} {cccccccc} \hline
& \multicolumn{2}{c}{Proposed Methods} & \multicolumn{5}{c}{Baselines} \\ \cmidrule(lr){2-3} \cmidrule(lr){4-8}
\text{Dataset} & Squared & Double Hinge & KMEANS & ITML & TL & SERAPH & SU\\ \hline
\text{adult} &62.72 (0.57) &59.74 (1.44) &71.44 (0.60) &71.79 (0.20) &58.53 (1.17) &70.54 (1.09) & \textbf{76.30 (0.04)}\\
\text{breast} & \textbf{96.90 (0.44)} & \textbf{96.53 (0.35)} & \textbf{96.28 (0.29)} & \textbf{96.79 (0.24)} &89.67 (1.97) & \textbf{96.68 (0.27)} &64.12 (0.91)\\
\text{diabetes} & \textbf{69.64 (0.68)} &67.08 (0.91) &66.27 (0.65) &64.87 (0.66) &63.15 (1.56) &67.44 (0.68) &33.90 (0.67)\\
\text{magic} &63.86 (1.44) & \textbf{70.37 (0.36)} &64.86 (0.15) &65.03 (0.13) &66.36 (0.30) &64.94 (0.14) &34.83 (0.15)\\
\text{phishing} & \textbf{75.52 (0.31)} &74.57 (0.37) &63.08 (0.47) &63.31 (0.41) &56.37 (1.18) &62.73 (0.76) &33.89 (0.20)\\
\text{spambase} &61.18 (1.11) &59.95 (1.38) & \textbf{63.55 (0.32)} & \textbf{64.17 (0.31)} &59.35 (1.48) & \textbf{63.53 (0.35)} &58.96 (0.44)\\
\text{mnist} & \textbf{74.23 (0.32)} & \textbf{75.19 (0.50)} &64.74 (0.55) &0.00 (0.00) &56.07 (0.87) &0.00 (0.00) &50.87 (0.26)\\
\text{fashion} &83.83 (0.55) & \textbf{87.86 (0.66)} &75.40 (0.34) &0.00 (0.00) &76.66 (1.39) &0.00 (0.00) &49.88 (0.08)\\
\text{cifar10} & \textbf{66.28 (1.77)} & \textbf{62.63 (2.53)} & \textbf{64.16 (0.01)} &0.00 (0.00) &61.26 (1.13) &0.00 (0.00) &59.05 (0.65)\\
\hline
\end{tabular}}
\end{table}

\begin{table}
\centering
\caption{Experimental results with class prior as $0.7$ and $200$ training triplets.}
\label{table:3}
\scalebox{0.73}{
\begin{tabular} {cccccccc} \hline
& \multicolumn{2}{c}{Proposed Methods} & \multicolumn{5}{c}{Baselines} \\ \cmidrule(lr){2-3} \cmidrule(lr){4-8}
\text{Dataset} & Squared & Double Hinge & KMEANS & ITML & TL & SERAPH & SU\\ \hline
\text{adult} &58.12 (0.90) &55.10 (1.00) &70.54 (1.50) &70.04 (1.17) &58.28 (0.94) &68.54 (1.67) & \textbf{75.27 (0.51)}\\
\text{breast} & \textbf{96.68 (0.32)} & \textbf{96.50 (0.35)} & \textbf{95.91 (0.34)} & \textbf{96.24 (0.24)} &94.27 (0.68) & \textbf{96.64 (0.28)} &66.20 (0.80)\\
\text{diabetes} & \textbf{69.25 (0.98)} &65.36 (0.89) &64.97 (0.87) & \textbf{67.27 (0.72)} &63.47 (1.22) & \textbf{67.11 (0.82)} &35.23 (0.94)\\
\text{magic} &60.54 (1.88) & \textbf{68.56 (0.53)} &64.88 (0.13) &65.15 (0.14) &66.31 (0.42) &64.97 (0.15) &34.60 (0.34)\\
\text{phishing} & \textbf{72.22 (0.62)} & \textbf{72.11 (0.65)} &63.70 (0.26) &63.71 (0.21) &57.02 (1.41) &63.17 (0.77) &34.03 (0.32)\\
\text{spambase} &57.69 (1.68) &55.74 (1.19) & \textbf{63.78 (0.34)} & \textbf{63.04 (0.35)} &60.78 (1.63) & \textbf{63.74 (0.25)} &58.92 (0.43)\\
\text{mnist} &67.14 (0.67) & \textbf{70.96 (0.53)} &64.49 (1.00) &0.00 (0.00) &57.88 (1.43) &0.00 (0.00) &50.10 (0.62)\\
\text{fashion} &76.67 (0.40) & \textbf{83.74 (0.55)} &74.90 (1.00) &0.00 (0.00) &73.24 (1.80) &0.00 (0.00) &47.97 (0.76)\\
\text{cifar10} & \textbf{63.14 (1.68)} &58.83 (2.16) & \textbf{64.16 (0.01)} &0.00 (0.00) &61.23 (1.18) &0.00 (0.00) &58.65 (0.66)\\
\hline
\end{tabular}}
\end{table}

\section{Conclusion}
In this paper, we proposed a novel method for learning a classifier from \textit{only} passively obtained triplet comparison data.
We established an estimation error bound for the proposed method, and confirmed that the estimation error decreases as the amount of triplet comparison data increases.
We also empirically confirmed that the performance of the proposed method surpassed multiple baseline methods on various datasets.
For future work, it would be interesting to investigate alternative methods that can handle a multi-class case.

\section*{Acknowledgments}
ZC was supported by the IST-RA program, the University of Tokyo.
NC was supported by MEXT scholarship.
IS was supported by JST CREST Grant Number JPMJCR17A1, Japan.
MS was supported by the International Research Center for Neurointelligence (WPI-IRCN) at The University of Tokyo Institutes for Advanced Study.
We thank Ikko Yamane and Han Bao for fruitful discussions on this work.

\bibliographystyle{APA}

\newpage 
\appendix
\section{Proof of Lemma 1}
\label{proof:lemma1}
\begin{proof}
From the data generation process,
we can consider the generation distribution for data of $\mathcal{D}_1$ as
\begin{equation}
  \label{eq:p1}
  \begin{split}
    p_1(\xs)
    & = p(\xs|(\ys)\in\mathcal{Y}_1) \\
    & = \frac{p(\xs, (\ys)\in\mathcal{Y}_1)}{p((\ys)\in\mathcal{Y}_1)} \\
    & = \frac{p(\xs,(\ys)\in\mathcal{Y}_1)}{\pip^3 + 2\pip^2\pin + 2\pip\pin^2 + \pin^3} .\\
  \end{split}
\end{equation}

Note that the denominator in Eq.\,\eqref{eq:p1} can be rewritten as
\begin{equation}
  \begin{split}
    \pit & \triangleq \pip^3 + 2\pip^2\pin + 2\pip\pin^2 + \pin^3 \\
    & = (\pip^3 + \pin^3) + 2(\pip^2\pin + \pip\pin^2) \\
    & = \pip^2 + \pip\pin + \pin^2 \\
    & = 1 - \pip\pin, \\
  \end{split}
\end{equation}

then we have
\begin{equation}
  \label{eq:p1simple}
   p_1(\xs) = \frac{p(\xs,(\ys)\in\mathcal{Y}_1)}{\pit}.
\end{equation}

Moreover, the distribution $p(\xs,(\ys)\in\mathcal{Y}_1)$ at the numerator of Eq.~\eqref{eq:p1simple} can be explicitly expressed as
\begin{equation}
  \begin{split}
    & \qquad p(\xs,(\ys)\in\mathcal{Y}_1) \\
    & = \pip^3\pp(x_a)\pp(x_b)\pp(x_c) + 
        \pip^2\pin\pp(x_a)\pp(x_b)\pn(x_c) +
        \pip\pin^2\pp(x_a)\pn(x_b)\pn(x_c) + \\
    & \quad \pip^2\pin\pn(x_a)\pp(x_b)\pp(x_c) + \pip\pin^2\pn(x_a)\pn(x_b)\pp(x_c)
    + \pin^3\pn(x_a)\pn(x_b)\pn(x_c), \\
  \end{split}
\end{equation}
from the assumption that three instances in each triplet comparison is generated independently.

Similarly, the underlying density for data of $\mathcal{D}_2$ can be expressed as
\begin{equation}
  \begin{split}
    p_2(\xs) & = p(\xs|(\ys)\in\mathcal{Y}_2) \\
    & = \frac{p(\xs, (\ys)\in\mathcal{Y}_2)}{p((\ys)\in\mathcal{Y}_2)} \\
    & = \frac{\pip^2\pin\pp(x_a)\pn(x_b)\pp(x_c) + \pip\pin^2\pn(x_a)\pp(x_b)\pn(x_c)}{\pip^2\pin+\pip\pin^2} \\
    & = \pip\pp(x_a)\pn(x_b)\pp(x_c) + \pin\pn(x_a)\pp(x_b)\pn(x_c).
  \end{split}
\end{equation}
\end{proof}

\section{Proof of Theorem 1}
\label{proof:1}
\begin{proof}
For simplicity, we give the proof of $\mathcal{D}_{2,a}$ and the other $5$ cases follow the similar proof. Noticing
\begin{equation}
 \mathcal{D}_2  \mathop{\sim}_{i.i.d.} p_2(\xs) = \pip\pp(x_a)\pn(x_b)\pp(x_c) + \pin\pn(x_a)\pp(x_b)\pn(x_c).
\end{equation}
In order to decompose the triplet comparison data distribution into pointwise distribution, we marginalize $p_2(\xs)$ with respect to $x_b$ and $x_c$:
\begin{equation}
\begin{split}
  & \int p_2(\xs) dx_bdx_c \\
  = \quad & \pip\pp(x_a)\int \pn(x_b)dx_b\int \pp(x_c)dx_c + \pin\pn(x_a)\int \pp(x_b)dx_b\int \pn(x_c)dx_c \\
  = \quad & \pip\pp(x_a)\int\frac{p(x_b,y=-1)}{p(y=-1)}dx_b\int\frac{p(x_c,y=+1)}{p(y=+1)}dx_c + \\
  & \qquad \pin\pn(x_a)\int\frac{p(x_b,y=+1)}{p(y=+1)}dx_b\int\frac{p(x_c,y=-1)}{p(y=-1)}dx_c \\
  = \quad & \pip\pp(x_a) + \pin\pn(x_a) \\
  = \quad & \tilde p_1(x_a) \\
\end{split}
\end{equation}
\end{proof}

\section{Proof of Lemma 2}
\label{proof:lemma2}
\begin{proof}
Notice that the equation has an infinite number of solutions.
Letting
\begin{equation}
  T \triangleq
  \begin{bmatrix}
  \pip&\pin\\A&B\\\pin&\pip
  \end{bmatrix},
\end{equation}
we resort to finding the Moore-Penrose pseudo inverse~\citep{Moore20, Penrose54}, which provides the minimum Euclidean norm solution to the above system of linear equations.

Let $T^*$ denote the conjugate transpose.
We have
\begin{equation}
  \begin{split}
    T^*T & =
    \begin{bmatrix}
    \pip^2+A^2+\pin^2 & 2\pip\pin+AB \\
    2\pip\pin+AB & \pin^2+B^2+\pip^2
    \end{bmatrix} =
    \begin{bmatrix}a&b\\b&c\end{bmatrix}.
  \end{split}
\end{equation}

In the next step, we need to take the inverse of the above $2\times2$ matrix.
To achieve a proper inverse matrix,
we need to introduce another assumption that $\pip\neq\frac12$, which guarantees $ac-b^2\neq0$.
Then
\begin{equation}
  (T^*T)^{-1} = \frac1{(ac-b^2)}
  \begin{bmatrix}c&-b\\-b&a\end{bmatrix}.
\end{equation}
Finally, the Moore-Penrose pseudo inverse is given by
\begin{equation}
  \begin{split}
    (T^*T)^{-1}T^*
    & = \frac1{(ac-b^2)}
    \begin{bmatrix}
    c\pip-b\pin & cA-bB & c\pin-b\pip \\
    -b\pip+a\pin & -bA+aB & -b\pin+a\pip
    \end{bmatrix}.
  \end{split}
\end{equation}

Thus we can express $\pp(x)$ and $\pn(x)$ in terms of $\tilde p_1(x)$, $\tilde p_2(x)$ and $\tilde p_3(x)$ as
\begin{equation}
\begin{split}
  \pp(x) & = \frac1{(ac-b^2)}\left((c\pip-b\pin)\tilde p_1(x)+(cA-bB)\tilde p_2(x)+(c\pin-b\pip)\tilde p_3(x)\right), \\
  \pn(x) & = \frac1{(ac-b^2)}\left((a\pin-b\pip)\tilde p_1(x)+(aB-bA)\tilde p_2(x)+(a\pip-b\pin)\tilde p_3(x)\right). \\
\end{split}
\end{equation}
\end{proof}

\section{Proof of Theorem 2}
\label{proof:2}
\begin{proof}
Using Equation \ref{eq:rewrite}, we can rewrite the classification risk as
\begin{equation}
\begin{split}
  R_\ell(f) & = \Ep_{p(x,y)}[\ell(f(x),y)] \\
  & = \pite\Ep_{\pp(x)}[\ell_+(x)] + (1-\pite)\Ep_{\pn(x)}[\ell_-(x)] \\
  & = \frac{\pite}{(ac-b^2)}\{(c\pip-b\pin)\Ep_{\tilde p_1(x)}[\ell_+(x)] + (cA-bB)\Ep_{\tilde p_2(x)}[\ell_+(x)] + (c\pin-b\pip)\Ep_{\tilde p_3(x)}[\ell_+(x)]\} + \\ & \qquad \frac{1-\pite}{(ac-b^2)}\{(a\pin-b\pip)\Ep_{\tilde p_1(x)}[\ell_-(x)]+(aB-bA)\Ep_{\tilde p_2(x)}[\ell_-(x)]+(a\pip-b\pin)\Ep_{\tilde p_3(x)}[\ell_-(x)]\},
\end{split}
\end{equation}
which can be then simplified as Equation \ref{eq:theorem2}.
\end{proof}

\section{Proof of Theorem 3}
\label{proof:3}
\begin{proof}
Letting
\begin{equation*}
\begin{split}
  C_1\triangleq\frac{\pite}{(c\pip-b\pin)(ac-b^2)}, \quad & C_2\triangleq\frac{1-\pite}{(a\pin-b\pip)(ac-b^2)}, \\
  C_3\triangleq\frac{\pite}{(cA-bB)(ac-b^2)}, \quad & C_4\triangleq\frac{(1-\pite)}{(aB-bA)(ac-b^2)}, \\
  C_5\triangleq\frac{\pite}{(c\pin-b\pip)(ac-b^2)}, \quad & C_6\triangleq\frac{(1-\pite)}{(a\pip-b\pin)(ac-b^2)},
\end{split}
\end{equation*}
and
\begin{equation}
 \begin{split}
  R_a(f) & = \Ep_{x\sim \tilde p_1(x)}[C_1\lfp+C_2\lfn], \\
  R_b(f) & = \Ep_{x\sim \tilde p_2(x)}[C_3\lfp+C_4\lfn], \\
  R_c(f) & = \Ep_{x\sim \tilde p_3(x)}[C_5\lfp+C_6\lfn],
 \end{split}
\end{equation}
we can simplify the unbiased risk estimator info the form
\begin{equation}
  R(f) = R_a(f) + R_b(f) + R_c(f).
\end{equation}

Then
$$ R(\hat f) - R(f^*) \leq 2\sup_{f\in\mathcal{F}}|R_a(f)-\hat R_a(f)| + 2\sup_{f\in\mathcal{F}}|R_b(f)-\hat R_b(f)| + 2\sup_{f\in\mathcal{F}}|R_c(f)-\hat R_c(f)|. $$
For the first term,
\begin{equation}
 \begin{split}
  \sup_{f\in\mathcal{F}}|R_a(f) - \hat R_a(f)| & = \sup_{f\in\mathcal{F}} \left| \Ep_{p_a(x)}[C_1\lfp+C_2\lfn] - \frac1n\sum_{i=1}^n\hat L \right| \\
  & \leq |C_1|\sup_{f\in\mathcal{F}}\left|\Ep_{p_a(x)}[\lfp] - \frac1n\sum_{i=1}^n\widehat\lfp\right| \\
  & \quad + |C_2|\sup_{f\in\mathcal{F}}\left|\Ep_{p_a(x)}[\lfn] - \frac1n\sum_{i=1}^n\widehat\lfn\right| \\
  & \leq |C_1|2\mathcal{R} + |C_1|\sqrt{\frac{C_\ell^2\log\frac2\delta}{2n}} + |C_2|2\mathcal{R} + |C_2|\sqrt{\frac{C_\ell^2\log\frac2\delta}{2n}} \\
  & = (|C_1|+|C_2|)\left(\frac{2\rho C_\mathcal{F}}{\sqrt{n}} + \sqrt{\frac{C_\ell^2\log\frac2\delta}{2n}} \right)
 \end{split}
\end{equation}
Combining three terms, Theorem 3 is proven.
\end{proof}

\section{CNN Structure for CIFAR10}
\label{cnnstructure}
The following structure is used:
\begin{itemize}
  \item Convolution ($3$ in/$32$ out-channels, kernel size $3$) with ReLU.
  \item Convolution ($32$ in/$32$ out-channels, kernel size $3$) with ReLU.
  \item Max-pooling (kernel size $2$, stride $2$).
  \item Repeat twice:
  \begin{itemize}
    \item Convolution ($32$ in/$32$ out-channels, kernel size $3$) with ReLU.
    \item Convolution ($32$ in/$32$ out-channels, kernel size $3$) with ReLU.
    \item Max-pooling (kernel size $2$, stride $2$).
  \end{itemize}
  \item Fully-connected ($512$ units) with ReLU.
  \item Fully-connected ($1$ unit).
\end{itemize}
\end{document}